\documentclass[11pt]{article}
\usepackage{PACLIC2021}
\usepackage{times}
\usepackage{latexsym}
\usepackage{graphicx}
\usepackage{amsmath}
\usepackage{multirow}
\usepackage{url}

\title{Autoencoding Language Model Based Ensemble Learning \\ for Commonsense Validation and Explanation}

\author{Ngo Quang Huy, Tu Minh Phuong, Ngo Xuan Bach\\
 Department of Computer Science,\\
 Posts and Telecommunications Institute of Technology, Hanoi, Vietnam\\  
 {\tt ngoquanghuy1999lp@gmail.com; \{phuongtm,bachnx\}@ptit.edu.vn}}

\date{}

\begin{document}
\maketitle

\begin{abstract}
 An ultimate goal of artificial intelligence is to build computer systems that can understand human languages. Understanding commonsense knowledge about the world expressed in text is one of the foundational and challenging problems to create such intelligent systems. As a step towards this goal, we present in this paper \textbf{ALMEn}, an \textbf{A}utoencoding \textbf{L}anguage \textbf{M}odel based \textbf{En}semble learning method for commonsense validation and explanation. By ensembling several advanced pre-trained language models including RoBERTa, DeBERTa, and ELECTRA with Siamese neural networks, our method can distinguish natural language statements that are against commonsense (validation subtask) and  correctly identify the reason for making against commonsense  (explanation selection subtask). Experimental results on the benchmark dataset of SemEval-2020 Task 4 show that our method outperforms state-of-the-art models, reaching 97.9\% and 95.4\% accuracies on the validation and explanation selection subtasks, respectively. 
\end{abstract}

\section{Introduction}
Natural language understanding (NLU) is a long-standing goal of artificial intelligence. To understand natural languages, computer systems need to have the ability to capture not only the semantic of text but also the commonsense knowledge expressed in the languages. Existing studies on commonsense understanding usually test computer systems whether they are equipped with commonsense knowledge in dealing with language processing tasks \cite{Cui:20,Mihaylov:18,Ostermann:18,Talmor:19,Wang:18,Zellers:18,Zhang:17}. Little work has been conducted to directly evaluate commonsense validation, where the lack of annotated datasets is one of the main reasons.  

To promote research in commonsense understanding and related areas, SemEval-2020, the International Workshop on Semantic Evaluation, devotes Task 4 for commonsense validation and explanation \cite{Wang:20}. The aim of the task is to directly test whether a computer system can distinguish natural language statements, that make sense from those that do not, and propose the suitable explanation. The task consists of three subtasks: validation, explanation selection, and explanation generation, in which the computer system need to identify the against-commonsense statement among similar ones, to choose the best explanation for the non-sensical\footnote{Words ``commonsense'' (against-commonsense) and ``sensical'' (non-sensical) are used interchangeably in this paper.} statement among several provided options, and to generate a reasonable explanation automatically, respectively. 

\begin{figure}[tbp]
	\begin{center}
		\includegraphics[width = 8.1cm]{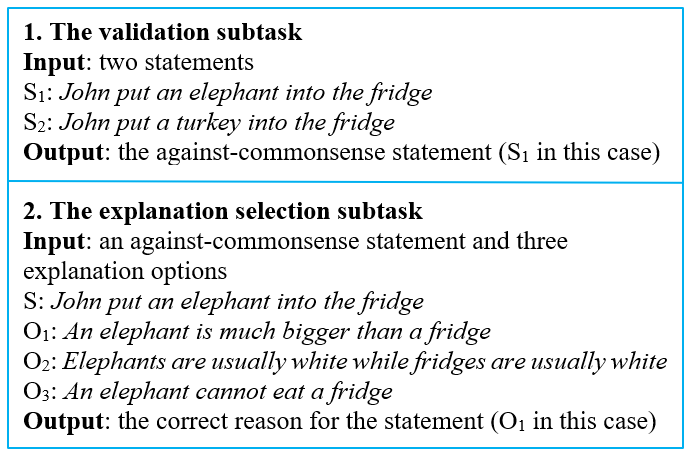}
	\end{center}	
	\vspace{-0.3cm}
	\caption{Commonsense validation and explanation selection subtasks \cite{Wang:20}.}
	\label{fig:examples}
\end{figure}

In this paper, we focus on the first two subtasks, i.e. the validation subtask and the explanation selection subtask. As illustrated in Figure \ref{fig:examples}, the input of the first subtask consists of two statements expressed in similar words. Among them, one statement is in commonsense and the other is not, and a system need to identify the against-commonsense one. In the second subtask, the system is provided with an against-commonsense statement and three explanation options, and the goal is to select the correct reason for the statement. 

Various methods have been introduced to deal with the commonsense validation and explanation selection subtasks. Most of them leverage Autoencoding Language Models (ALMs) pre-trained on a huge amount of plain text to capture knowledge about the world expressed in the text \cite{Liu:20,Srivastava:20,Wang:20,Xing:20,Zhang:20,Zhao:20}. The advantage of ALMs is that it can model bidirectional contexts by reconstructing the original text from the corrupted input ([MASK]) in some way. With the availability of pre-trained language models, this approach can directly develop models for commonsense validation and explanation from annotated training datasets. To improve the performance, some works proposed to exploit external knowledge bases, which contain explicit and structured knowledge \cite{Zhang:20,Zhao:20}, in addition to unstructured text.   

Here, we present an ensemble learning method for commonsense validation and explanation selection subtasks. Our key idea is to build multiple models using different pre-trained ALMs, which are then trained further on a closely related task. Each single model employs a Siamese neural network, which uses the same weights while working simultaneously on two or three different input vectors to compute comparable output vectors. The final decision is the weighted combination of single systems' outputs where the weights are optimized using differential evolution search. We empirically verify the effectiveness of the proposed method on the benchmark dataset of SemEval-2020 Task 4. Experimental results show that our method outperforms existing models including methods that utilize external knowledge bases.  

The rest of this paper is organized as follows. Section \ref{sec:related} describes related work. Section \ref{sec:method} introduces our proposed method for commonsense validation and explanation. Experimental setup and results are presented in Sections \ref{sec:setup} and \ref{sec:result}, respectively. Finally, Section \ref{sec:conclusion} concludes the paper and discusses future work.                          

\section{Related Work}
\label{sec:related}
This section describes related work on autoencoding language models and existing methods for commonsense validation and explanation selection. 

\textbf{Autoencoding Language Models (ALMs)}. 
Unlike autoregressive language models which aim to predict data from the past input (unidirectional), autoencoding language models can capture both left and right contexts by reconstructing the original sentence from the corrupted input. ALMs are pre-trained on a huge amount of plain text to build bidirectional representations of the whole sentences. They can be fine-tuned and achieve state-of-the-art results on many natural language processing tasks, including sentence classification, token classification, text generation, and question answering \cite{Clark:20,Devlin:19,He:21,Liu:19}. Here, we give a brief review of notable ALMs, including BERT, RoBERTa, DeBERTa, and ELECTRA.  

Introduced by Devlin et al. \cite{Devlin:19}, BERT (\underline{B}idirectional \underline{E}ncoder \underline{R}epresentations from \underline{T}ransformers) is a typical ALM. The key idea of BERT is using a deep bidirectional Transformer \cite{Vaswani:17}, which allows the model to fuse the left and the right context of token representations. The model is pre-trained with two objectives: 1) the masked language model objective \cite{Taylor:53}, which reconstructs the original vocabulary of the masked tokens from the input; and 2) the next sentence prediction task, which predicts whether a sentence B is actually the next sentence that follows a sentence A in a monolingual corpus. RoBERTa (\underline{R}obustly \underline{o}ptimized \underline{BERT} \underline{a}pproach), proposed by Liu et al. \cite{Liu:19}, uses the same architecture of BERT but it modifies hyper-parameters, removes the next sentence prediction pre-training objective, and trains with much larger mini-batches and learning rates. Besides, there are other modifications such as using dynamic masking (tokens are masked differently at each epoch) and a larger byte-level with byte-pair encoding \cite{Sennrich:16}. Following the success of BERT and RoBERTa, He et al. \cite{He:21} present DeBERTa (\underline{De}coding-enhanced \underline{BERT} with disentangled \underline{a}ttention), that improves the BERT and RoBERTa models using a disentangled attention mechanism and an enhanced mask decoder. Unlike BERT, RoBERTa, and DeBERTa which are masked language models, ELECTRA (\underline{E}fficiently \underline{L}earning an \underline{E}ncoder that \underline{C}lassifies \underline{T}oken \underline{R}eplacements \underline{A}ccurately) \cite{Clark:20} uses a more sample-efficient pre-training task called replaced token detection. The model corrupts the input by replacing some tokens with plausible alternatives sampled from a small generator network, and the pre-training objective is predicting whether each token in the corrupted input was replaced by a generator sample or not  

\textbf{Methods for Commonsense Validation and Explanation}.
Existing methods for commonsense validation and explanation selection usually utilize ALMs such as BERT \cite{Devlin:19}, RoBERTa \cite{Liu:19}, and ALBERT \cite{Lan:20}, which are adopted by fine-tuning on the training set of two subtasks. Xing et al. \cite{Xing:20} with IIE-NLP-NUT, Srivastava et al. \cite{Srivastava:20} with Solomon, Liu et al. \cite{Liu:20} with LMVE, Saeedi et al. \cite{Saeedi:20} with CS-NLP, Dash et al. \cite{Dash:20} with CS-NET, Doxolodeo and Mahendra \cite{Doxolodeo:20} with UI,  and Teo \cite{Teo:20} with TR, are examples of such methods, among the others. Some of them are intermediate pre-trained on Natural Language Inference (NLI) datasets before fine-tuning. Zhang et al. \cite{Zhang:20} and Zhao et al. \cite{Zhao:20} propose to exploit external structured knowledge bases in addition to unstructured text. Their systems, CN-HIT-IT.NLP and ECNU-SenseMaker, achieve the best results on the dataset of SemEval-2020 Task 4. While the former system enhances the language representations of the text by injecting relevant triples from the ConceptNet \cite{Speer:17} to build a knowledge-rich sentence tree, the later one integrates knowledge from ConceptNet using a knowledge-enhanced graph attention network. 

Perhaps the most closely related work to ours are the methods of Liu \cite{LiuQiaoNing:20} and Mohammed and Abdullah \cite{Mohammed:20}. Both of them use ensemble learning of pre-trained language models for commonsense validation and explanation. The proposed method in this paper, however, is different from those methods in some aspects. First, we utilize Siamese networks, which use the same weights while working simultaneously on multiple different input vectors, and therefore, are efficient to compare similar vectors. Second, we carefully select single models to ensemble, which have different characteristics and will be shown to be non-complete-overlapping in experiments. Finally, we use differential evolution search to optimize the weights of the ensemble model instead of using simple voting techniques. Experimental results show that our model outperforms those methods by large margins.    

\section{Method}
\label{sec:method}

\subsection{Method Overview}
\begin{figure*}[tbp]
	\begin{center}
		\includegraphics[width = 15cm]{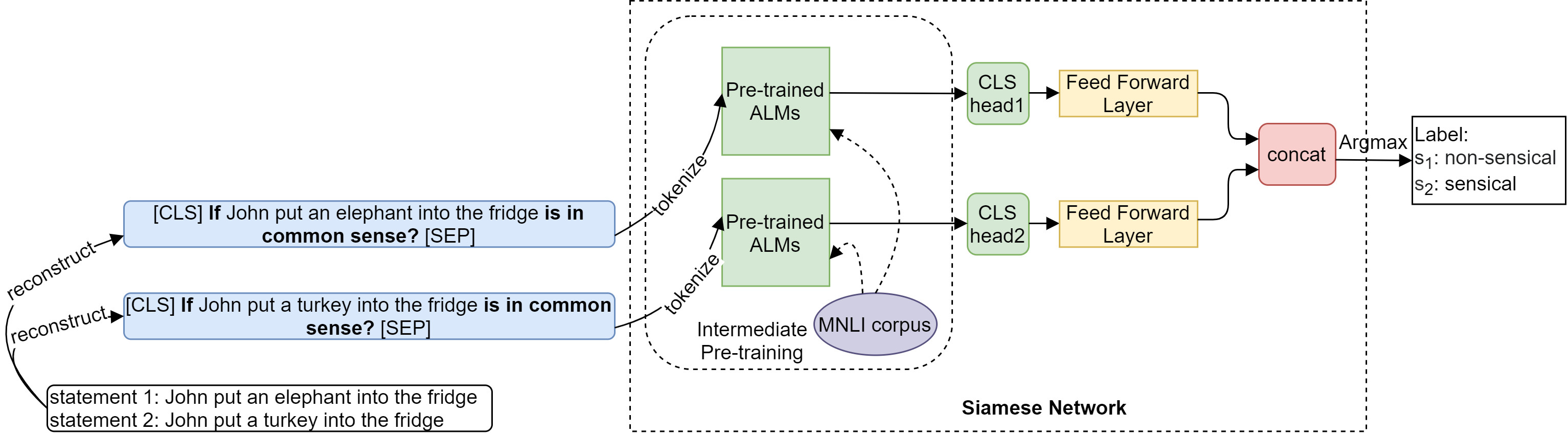}
	\end{center}	
	\vspace{-0.3cm}
	\caption{The architecture of single models.}
	\label{fig:single}
\end{figure*}

The general idea of our approach is to build an ensemble model from several single ones. The single models share the same architecture but use different autoencoding language models (ALMs). Figure \ref{fig:single} shows the architecture of a single model for the commonsense validation subtask with three main components: input reconstruction, autoencoding language models with intermediate pre-training, and Siamese network.

\begin{itemize}
	\item \textbf{Input reconstruction}: input statements are first reconstructed by adding more information and special tokens. 
	\item \textbf{ALMs with intermediate pre-training}: modified statements are then fed into ALMs, which have been pre-trained on a huge amount of raw text and tuned (intermediate pre-trained) on Multi-genre Natural Language Inference (MNLI) dataset \cite{Williams:18}.  
	\item \textbf{Siamese network}: ALMs are integrated into a Siamese network, which contains two parameter-sharing similar subnets corresponding to two input statements. The output of the Siamese network is used to decide which statement is sensical, and therefore, which one is non-sensical. 
\end{itemize} 

Note that the single models for the explanation selection subtask also share this architecture except that they use three inputs for explanation candidates instead of two inputs for statements. 

\subsection{Input Reconstruction}
Instead of just tokenizing each input sentence (statement or explanation option) and adding special tokens [CLS] and [SEP] at the very beginning and at the end of it, we reconstruct the sentence by adding additional context. Additional context, which can be thought of as a task descriptor given to the model, can define the task, help the language models to extract more detailed information of the commonsense so the model could understand the semantics of input sequences. This type of trick has been utilized widely since the introduction of BERT \cite{Devlin:19}. For commonsense validation and explanation selection, it is also used in IIE-NLP-NUT of Xing et al. \cite{Xing:20}. 

In the first subtask, our system is required to find out the statement that is against commonsense from two given statements. We propose transforming an input statement $S_i$ into the following retrieval based format: [CLS] \emph{If $S_i$ is in common sense?} [SEP], where $i \in \{1, 2\}$. For example, the statement ``\emph{John put an elephant into the fridge}'' will become [CLS] \emph{If John put an elephant into the fridge is in common sense?} [SEP]. 

In the second subtask, the mission is to select the most reasonable explanation from three given options to interpret for a false statement. We propose concatenating the false statement with each explanation option by a connector so that the final format of the input sequence like: [CLS] \emph{$S_f$  does not make sense because $O_j$} [SEP], where $S_f$ denotes the false statement, $O_j$ is an option, and $j \in \{1, 2, 3\}$. For example, with the false statement ``\emph{John put an elephant into the fridge}'' and the explanation option ``\emph{An elephant is much bigger than a fridge}'', we will have the input as follows: [CLS] \emph{John put an elephant into the fridge does not make sense because An elephant is much bigger than a fridge} [SEP].

\subsection{ALMs with Intermediate Pre-training}
Any autoencoding language model (ALM) can be used in our single models. In this work, to create an effective ensemble model, we build three single models with RoBERTa, DeBERTa, and ELECTRA. RoBERTa \cite{Liu:19} is a good choice for NLU tasks because it is built on BERT with some contributions to the amount of data and removing the next sentence prediction pre-training objective, which helps the model focus on language understanding. In fact, RoBERTa has been chosen in several systems for commonsense validation and explanation \cite{Srivastava:20,Xing:20}. DeBERTa \cite{He:21} gives a solution for the disadvantage of self-attention mechanism (lacking a natural way to encode word position information) by using two vectors, which encode content and position respectively, instead of one vector that is the sum of word embedding and position embedding in BERT architecture.  At the time of conducting this research, DeBERTa achieves the best results on many NLU tasks \cite{He:21}. Unlike BERT-based architectures such as RoBERTa and DeBERTa, which are pre-trained with the Masked Language Model (MLM) objective, ELECTRA \cite{Clark:20} uses a different approach. Instead of masking a subset of tokens in the input, ELECTRA corrupts it by replacing some tokens with plausible alternatives sampled from a generator network. A discriminator network is then trained to predict whether each token in the corrupted input was replaced or not. This kind of pre-training is more efficient than the MLM objective because the task can be defined over all input tokens.

ALMs are pre-trained on massive, open-domain text corpora to understand the language and to capture world knowledge. Usually, a pre-trained language  model is directly fine-tuned on training datasets to perform NLP classification tasks. A method to improve the classification performance is training the language model on an intermediate task before fine-tuning it on the downstream target task. So before adopting pre-trained ALMs on commonsense validation and explanation selection, we conduct intermediate pre-training on the commonsense corpus MNLI \cite{Williams:18}, which helps our model understands both the language and commonsense better.

\subsection{Siamese Network}
Let $(\textbf{s}_1, \textbf{s}_2)$ denote a pair of input statements. As illustrated in Figure \ref{fig:single}, the Siamese network computes an output label as follows\footnote{For notation, we denote vectors with bold lower-case, matrices with bold upper-case, and scalars with italic lower-case.}: 
$$\textbf{c}_i = \text{ALM}(\textbf{s}_i), score_i = \text{FFN}(\textbf{c}_i),  \nonumber$$
$$\textbf{x} = (score_1, score_2), label = \text{Argmax}(\textbf{x}), \nonumber$$
where $\textbf{c}_i$ is the representations of the [CLS] token and $score_i$ denotes the estimated sensical score for statement $\textbf{s}_i$ $(i=1,2)$; ALM$(\cdot)$ is the pre-trained autoencoding language model; FFN$(\cdot)$ indicates the feed forward network layer; and $label$ is the predicted index of the sensical statement. 

Let $y$ denote the true index of the sensical statement, we define the loss functions for a single training sample and for a batch of samples as follows:
\begin{align*}
\mathcal{L}(\textbf{x}, y) &= - \log(\frac{\exp(\textbf{x}[y])}{\sum_{i}^{}\exp(\textbf{x}[i])}) \\
&= - \textbf{x}[y] + \log(\sum_{i}^{}\exp(\textbf{x}[i])), \text{and} 
\end{align*}

$$\mathcal{L}(\textbf{X}, \textbf{y}) = \frac{1}{N}\sum_{j=1}^{N}{\mathcal{L}(\textbf{X}[j], \textbf{y}[j])}, $$
where $N$ denotes the batch size and $\textbf{X}[j]$ and $\textbf{y}[j]$ are the concatenation vector of the predicted sensical scores and the true index of the $j^{th}$ sample. The parameters of single models are updated for each training batch.

\subsection{Ensemble Learning}
\begin{figure*}[tbp]
	\begin{center}
		\includegraphics[width = 13cm]{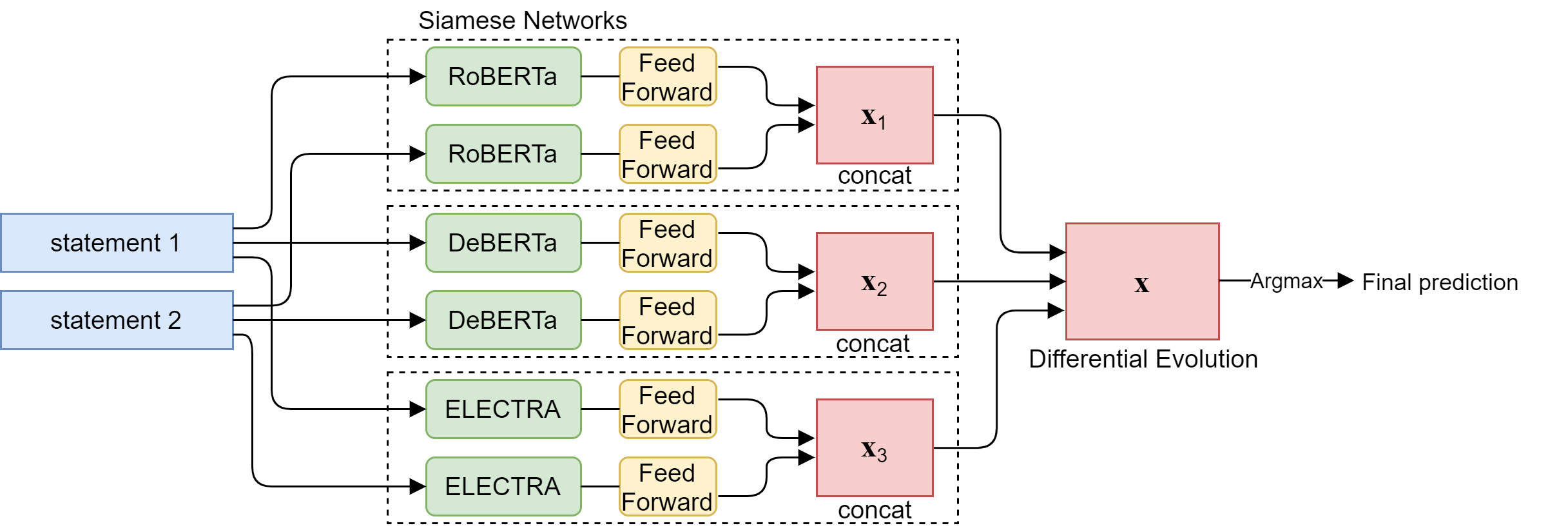}
	\end{center}	
	\vspace{-0.3cm}
	\caption{Our proposed ensemble learning model.}
	\label{fig:ensemble}
\end{figure*}

Figure \ref{fig:ensemble} illustrates our ensemble learning model for the commonsense validation subtask. Two (reconstructed) input statements are fed into three single models, which use a Siamese network consisting of an ALM with intermediate pre-training followed by a feed forward network. The outputs of single models are then ensembled to produce the final prediction. 

Let $\textbf{x}_1$, $\textbf{x}_2$, and $\textbf{x}_3$ denote the outputs (concatenation vectors) of single models using RoBERTa, DeBERTa, and ELECTRA, respectively, the output $\textbf{x}$ of the ensemble model can be computed as follows:
\begin{equation}
\textbf{x}= w_1 \textbf{x}_1 + w_2 \textbf{x}_2 + w_3 \textbf{x}_3, \nonumber
\end{equation}
where $w_i$ $(i=1,2,3)$ are real numbers (weights) which measure the contribution of single models. Recall that $\textbf{x}_i$ and $\textbf{x}$ are two-dimensional vectors consisting of two real numbers which estimate the sensical scores of input statements. The commonsense statement is then picked up through an Argmax operation, and therefore, the other statement is against-commonsense. To find the best values of weights $w_i$ $(i=1,2,3)$, we utilize differential evolution search \cite{Storn:97}, a stochastic population based method for global optimization problems. In our experiments $w_i$ $(i=1,2,3)$ were optimized using the development set.  

The ensemble model for the explanation selection subtask is the same as the model shown in Figure \ref{fig:ensemble} except that it has three inputs for explanation options instead of two inputs for statements.

\section{Experimental Setup}
\label{sec:setup}

\subsection{Data and Evaluation Method}	
We conducted experiments using the benchmark dataset of SemEval-2020 Task 4. As shown in Table \ref{tab:dataset}, the dataset consists of 11,997 data samples, divided into training/development/test subsets with 10,000/997/1,000 samples, respectively. Each sample contains a pair of sensical and non-sensical statements (for the validation subtask) and three reasons which explain why the non-sensical statement does not make sense (for the explanation selection subtask). One of the three reasons is correct while the others are confusing. The average length of different types of sentences in the dataset can also be found in Table \ref{tab:dataset}. 

\begin{table}[t]
	\centering
	\caption{Statistical information of SemEval-2020 Task 4 dataset \cite{Wang:20}}
		\label{tab:dataset}
		\begin{tabular}{l|c|c|c}
			\hline	
			\hline
				& \textbf{Train} &	\textbf{Dev} &	\textbf{Test}\\
				\hline
				\hline
				Number of samples	& 10,000 &	997 &	1,000\\
			\hline
			\hline
			\multicolumn{4}{c}{\textbf{Average length}}\\
			\hline
			\hline			
			Sensical statements &	7.67 & 7.12 &	7.25\\
			\hline
			Non-sensical Statements &	7.69 & 7.16 &	7.36\\
			\hline
			Correct Reasons &	8.13 & 7.96 & 8.09\\
			\hline
			Confusing Reasons &	7.80 & 7.14 &	7.29\\
			\hline
			\hline
			\end{tabular}	
\end{table}

For both the validation and explanation selection subtasks, systems were evaluated using accuracy on the test set. The training and development sets were used to train the systems and tune parameters, respectively. 

\subsection{Network Training}
Our models were implemented in PyTorch\footnote{\url{https://pytorch.org/}} using Huggingface's Transformers\footnote{\url{https://huggingface.co/transformers/}}, an open source library that provides thousands of pre-trained models to perform NLU/NLG tasks. Along with that, we also used Differential Evolution API provided by SciPy\footnote{\url{https://www.scipy.org/}}, an open source library for scientific computing. We set the batch size to $32, 8, 24$ for RoBERTa, DeBERTa, and ELECTA, respectively as they have varying number of parameters. The max sequence length was tuned in $[30,40,50,60,80,100]$ for all single models and the best value was $50$ for both subtasks. Our models were trained using the AdamW optimizer \cite{Loshchilov:19}, a stochastic optimization method that modifies the typical implementation of weight decay in the Adam optimizer \cite{Kingma:15} by decoupling weight decay from the gradient update. The epsilon and weight decay were set to default values in PyTorch, i.e. 1$e$-8 and 1$e$-2, respectively. The learning rate was tuned in [1$e$-5, 1.5$e$-5, 2$e$-5, 3$e$-5, 5$e$-5, 1$e$-4] with a linear warm up schedule in which the number of warm up steps was equal to 10\% of the number of training steps. To reduce overfitting, we also added dropout with a dropout rate of $0.1$ to all hidden layers. We trained our models for $10$ epochs and tuned on the development set. The final values of the hyperparameters are shown in Table \ref{tab:params}.

\begin{table}[t]
	\centering
	\caption{Final values of the hyperparameters}
		\label{tab:params}
		\begin{tabular}{l|l}
			\hline	
			\hline
				\textbf{Hyperparameter} &	\textbf{Value}\\
			\hline
			\hline
				\multirow{3}{*}{Batch sizes} & 32 (RoBERTa)\\
						& 8 \ (DeBERTa)\\
						&	24 (ELECTRA)\\
			  \hline
				Maximal sequence length &	50\\
				\hline
				Dropout rate &	0.1\\
				\hline
				\#epochs &	10\\
				\hline
				Learning rate &	1$e$-4\\
				\hline
				Weight decay coefficient & 	1$e$-2\\
				\hline
				Bounds of search space &	[0.0:1.0]\\
				\hline
				\#iterations of DE search &	10,000\\
				\hline
				Relative tolerance & 	1$e$-7\\
			\hline
			\hline
			\end{tabular}	
\end{table}

\subsection{Models to Compare}
We conducted experiments to compare our proposed models (ensemble model and three single ones with RoBERTa\footnote{We used the pre-trained model of Facebook. Available at: \url{https://github.com/pytorch/fairseq/blob/master/examples/roberta}}, DeBERTa\footnote{We used the pre-trained model of Microsoft. Available at: \url{https://github.com/microsoft/DeBERTa}}, and ELECTRA\footnote{We used the pre-trained model of Hugging Face. Available at: \url{https://huggingface.co/ynie/electra-large-discriminator-snli_mnli_fever_anli_R1_R2_R3-nli}}, respectively) and existing models (four best models on the dataset of SemEval-2020 Task 4, including CN-HIT-IT.NLP \cite{Zhang:20}, ECNU \cite{Zhao:20}, IIE-NLP-NUT \cite{Xing:20}, and Solomon \cite{Srivastava:20}; two ensemble models QiaoNing \cite{LiuQiaoNing:20} and JUST \cite{Mohammed:20}). Table \ref{tab:models} summarizes the models and their properties.
\begin{table*}[tbp]
	\centering
	\caption{Our proposed models and existing models to be compared}
		\label{tab:models}
		\begin{tabular}{c|l|c|l}
			\hline	
			\hline
			\textbf{\#} &	\textbf{Model} &	\textbf{Pre-trained LMs} & \textbf{Properties}\\
			\hline			
			\hline
			\multicolumn{4}{c}{\textbf{Our proposed models}} \\
			\hline
			\hline
			1	& Single-RoBERTa &	RoBERTa &	- Input reconstruction \\
			\cline{1-3}
			2	& Single-DeBERTa &	DeBERTa	& - Intermediate pre-training\\
			\cline{1-3}
			3	& Single-ELECTRA &	ELECTRA	 & - Siamese network\\
			\hline
			4	& ALMEn &	\begin{tabular}{@{}l@{}}Ensemble (RoBERTa, \\ DeBERTa, ELECTRA) \end{tabular} & 	- Differential evolution search \\
			\hline
			\hline
			\multicolumn{4}{c}{\textbf{Existing models}} \\
			\hline
			\hline		
			5	& CN-HIT-IT.NLP & K-BERT \cite{Liu:20KBERT} &	- Knowledge from ConceptNet \\
			\hline	
			\multirow{2}{*}{6}	& \multirow{2}{*}{ECNU} & &	- Knowledge from ConceptNet \\
			& & & - Graph attention network\\
			\hline
			\multirow{2}{*}{7}	& \multirow{2}{*}{IIE-NLP-NUT }	& \multirow{2}{*}{RoBERTa} &	- Input reconstruction \\
			& & & - Intermediate pre-training \\
			\hline
			8	& Solomon & RoBERTa &	\\
			\hline
			9 &	QiaoNing & \begin{tabular}{@{}l@{}}Ensemble (RoBERTa,\\ BERT, ALBERT, XLNet) \end{tabular} &	- Weighted sum \\
			\hline
			10 &	JUST & \begin{tabular}{@{}l@{}}Ensemble (RoBERTa,\\ BERT, ALBERT, XLNet) \end{tabular} &	- Voting \\
			\hline
			\hline
			\end{tabular}	
\end{table*}

\section{Experimental Results}
\label{sec:result}

\subsection{Ensemble Model vs. Single Models}
We first conducted experiments to compare our ensemble model (ALMEn) with the single models. The purpose of these experiments are two-fold: (1) to investigate the performance of the single models with different ALMs; and (2) to evaluate the performance of the ensemble model in comparison with the single models. Tables \ref{tab:ResultProposedSubtaskA} and \ref{tab:ResultProposedSubtaskB} show experimental results of our models on the commonsense validation and explanation selection subtasks, respectively. For both subtasks, the model with DeBERTa achieved the best results among the three single ones, 96.7\% on validation and 94.2\% on explanation selection. The above results are reasonable because DeBERTa is one of the latest models which has been shown to be better than other ALMs including RoBERTa and ELECTRA on many NLP tasks \cite{He:21}. As expected, ALMEn outperformed all three single models by large margins. ALMEn achieved 97.9\% accuracy on the validation subtask and 95.4\% accuracy on the explanation selection subtask, which improved 1.2\% compared with the best single models in both subtasks.   
\begin{table}[t]
	\centering
	\caption{Results of the proposed models on the validation subtask}
		\label{tab:ResultProposedSubtaskA}
		\begin{tabular}{c|l|c}
			\hline	
			\hline
			\#	& \textbf{Model} &	\textbf{Accuracy(\%)}\\
			\hline
			\hline
			1	& Single-RoBERTa & 	95.9\\
			\hline
			2	& Single-DeBERTa & 	96.7\\
			\hline
			3	& Single-ELECTRA & 	92.6\\
			\hline
			\textbf{4}	& \textbf{ALMEn} & 	\textbf{97.9}\\
			\hline
			\hline
				& Human	& 99.1\\
			\hline			
			\hline		
		\end{tabular}	
\end{table}

\begin{table}[t]
	\centering
	\caption{Results of the proposed models on the explanation selection subtask}
		\label{tab:ResultProposedSubtaskB}
		\begin{tabular}{c|l|c}
			\hline	
			\hline
			\#	& \textbf{Model} &	\textbf{Accuracy(\%)}\\
			\hline
			\hline
			1	& Single-RoBERTa & 	92.4\\
			\hline
			2	& Single-DeBERTa & 	94.2\\
			\hline
			3	& Single-ELECTRA & 	92.8\\
			\hline
			\textbf{4}	& \textbf{ALMEn} & \textbf{95.4}\\
			\hline
			\hline
				& Human	& 97.8\\
			\hline			
			\hline		
		\end{tabular}	
\end{table}

\subsection{Ensemble Model vs. Existing Systems}	
Next, we compared our ensemble model with existing systems on commonsense validation and explanation selection subtasks. For existing systems, we used the results published in their papers, which used the same training/development/test split of the SemEval-2020 Task 4 dataset. As showed in Tables \ref{tab:ResultExistingSubtaskA} and \ref{tab:ResultExistingSubtaskB}, ALMEn outperformed the best existing models on both subtasks. For the first subtask, ALMEn improved 0.9\% and 1.2\% compared with the best models CN-HIT-IT.NLP and ECNU-SenseMaker, respectively. For the second one, ALMEn also improved 0.4\% and 0.6\% in comparison with ECNU-SenseMaker and CN-HIT-IT.NLP. We also note that both CN-HIT-IT.NLP and ECNU-SenseMaker exploits external knowledge sources, i.e. ConceptNet, in addition to pre-trained language models. Compared with existing ensemble models, i.e. QiaoNing and JUST\footnote{JUST only shows the result on the validation subtask.}, our model even outperformed by larger margins, 2.0\% in the first subtask and 4.6\% in the second one.   
\begin{table}[t]
	\centering
	\caption{ALMEn vs. existing systems on the validation subtask (KB stands for knowledge base and En means ensemble learning)}
		\label{tab:ResultExistingSubtaskA}
		\begin{tabular}{c|l|c|c|c}
			\hline	
			\hline
				\#	& \textbf{Model} &	\textbf{Acc(\%)} &	\textbf{KB?} &	\textbf{En?} \\
				\hline
				\hline
				    & Human	& 99.1 & & \\
				\hline
				\hline
				\textbf{4} &	\textbf{(Our) ALMEn} & \textbf{97.9} &	No &	\textbf{Yes}\\
				\hline
				\hline
				5	& CN-HIT-IT.NLP  &	97.0 &	\textbf{Yes} &	No\\
				\hline
				6	& ECNU & 96.7 &	\textbf{Yes} &	No\\
				\hline
				7	& IIE-NLP-NUT &	96.4 &	No &	No\\
				\hline
				8	& Solomon & 	96.0 &	No &	No\\
				\hline
				9	& QiaoNing &	95.9 &	No &	\textbf{Yes}\\
				\hline
				10 &	JUST &	89.1 &	No &	\textbf{Yes}\\
			\hline			
			\hline		
		\end{tabular}	
\end{table}

\begin{table}[t]
	\centering
	\caption{ALMEn vs. existing systems on the explanation selection subtask (KB stands for knowledge base and En means ensemble learning)}
		\label{tab:ResultExistingSubtaskB}
		\begin{tabular}{c|l|c|c|c}
			\hline	
			\hline
				\#	& \textbf{Model} &	\textbf{Acc(\%)} &	\textbf{KB?} &	\textbf{En?} \\
				\hline
				\hline
				    & Human	& 97.8 & & \\
				\hline
				\hline
				\textbf{4} &	\textbf{(Our) ALMEn} & \textbf{95.4} &	No &	\textbf{Yes}\\
				\hline
				\hline
				6	& ECNU &	95.0 &	\textbf{Yes} &	No\\
				\hline
				5	& CN-HIT-IT.NLP &	94.8 &	\textbf{Yes} &	No\\
				\hline
				7	& IIE-NLP-NUT &	94.3 &	No &	No\\
				\hline
				8	& Solomon & 94.0 &	No &	No\\
				\hline
				9	& QiaoNing &	90.8 &	No &	\textbf{Yes}\\
				\hline
				10 &	JUST &	- &	- &	\textbf{Yes}\\
			\hline			
			\hline		
		\end{tabular}	
\end{table}

\subsection{Contribution of Single Models}
\begin{figure*}[tbp]
	\begin{center}
		\includegraphics[width = 11cm]{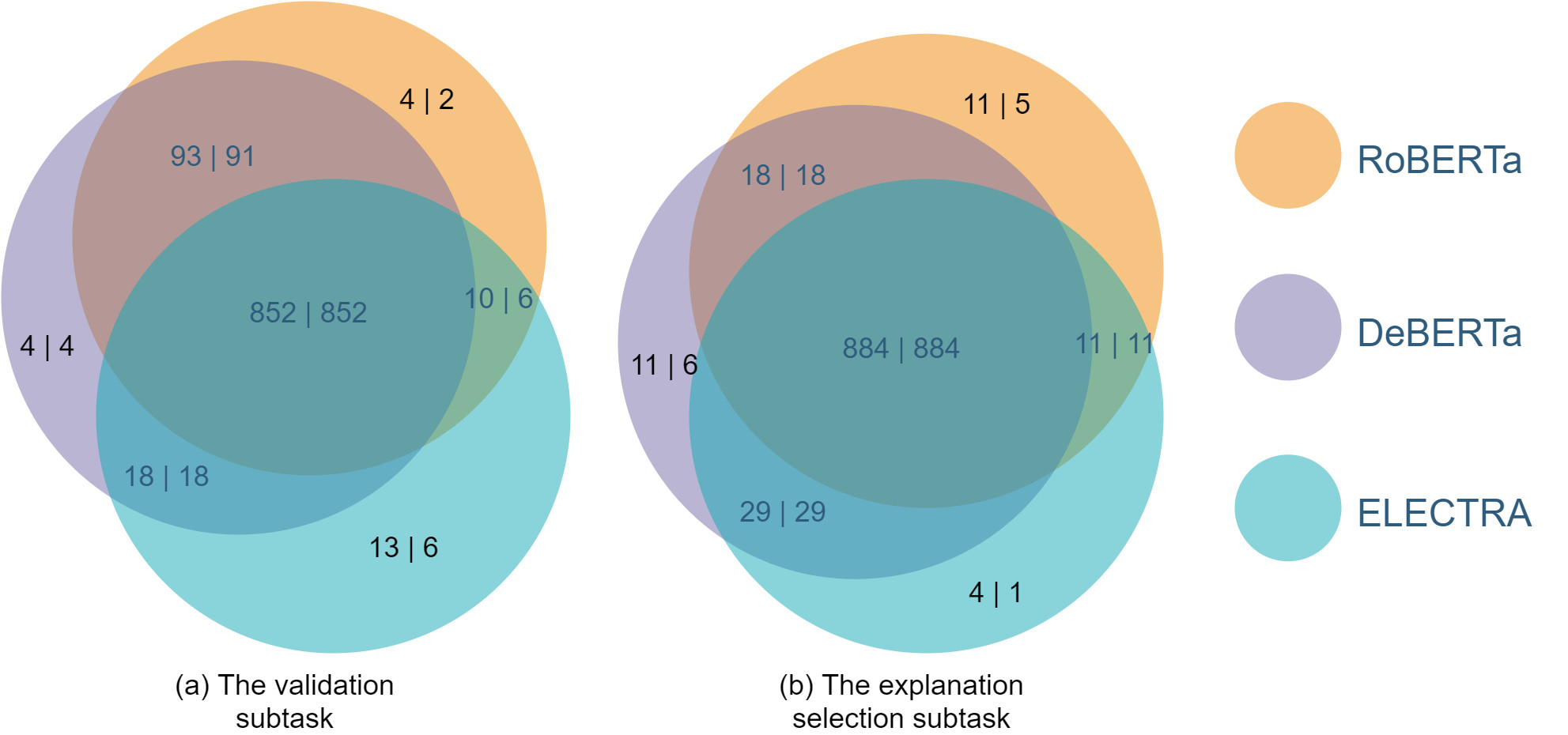}
	\end{center}	
	\vspace{-0.3cm}
	\caption{The overlap of single models on the test set.}
	\label{fig:overlap}
\end{figure*}

We also investigate the contribution of the single models to the success of ALMEn by looking more closely on the overlap of their results on the test set. Figure \ref{fig:overlap} shows Venn diagrams illustrating the overlap of the single models on two subtasks. Each section of the diagram for a subtask contains two numbers in the format $\alpha|\beta$, where $\alpha$ denotes the number of the test samples correctly classified by the only related single models, and $\beta$ is the number of the test samples among them correctly predicted by ALMEn. For example, $4|2$ on the left diagram means that there were 4 test samples correctly classified by RoBERTa model (incorrectly classified by DeBERTa and ELECTRA models), and only 2 of them were predicted correctly by ALMEn. $93|91$ on the left diagram indicates that there were 93 test samples correctly classified by RoBERTa and DeBERTa models (incorrectly classified by ELECTRA model), and only 91 of them were predicted correctly by ALMEn. 
   
Let $A$, $B$, and $C$ denote the subsets of the test samples that were classified accurately by all the three single models, only two single models, and only one single model, respectively. It is straightforward to see that in both subtasks ALMEn gave correct answers for all samples in $A$. The next observation is that ALMEn predicted correctly almost all of samples in $B$ (115 of 121 for validation; 58 of 58 for explanation selection). These results show that our ensemble model also reflects the majority voting strategy, a key principle of ensemble learning. Furthermore, ALMEn also dealt well with many samples in $C$ (12 of 21 for validation; 12 of 26 for explanation selection). The fact that all three models contributed correct samples in C proves their important role in the ensemble model. 

\section{Conclusion}
\label{sec:conclusion}
We have presented in this paper an ensemble learning method for commonsense validation and explanation selection. By utilizing multiple autoencoding language models with Siamese neural networks, our method outperformed state-of-the-art models on the benchmark dataset of SemEval-2020 Task 4. Without using any external knowledge source, our method is easy to adapt to other languages. We plan to investigate autoencoding language model based ensemble learning for other natural language processing tasks, including sentence classification and sequence labeling, among the others. 

\end{document}